%% file: touch_iros.tex
\documentclass[letterpaper, 10 pt, conference]{ieeeconf}  
\IEEEoverridecommandlockouts                              
\input{packages}
\input{macros}

\title{\LARGE \bf
3D Shape Perception from Monocular Vision, Touch, and Shape Priors
}

\author{Shaoxiong Wang$^{*1}$, Jiajun Wu$^{*1}$, Xingyuan Sun$^{1,2}$, Wenzhen Yuan$^{1}$,\\ William T. Freeman$^{1,3}$, Joshua B. Tenenbaum$^{1}$, and Edward H. Adelson$^{1}$
\thanks{$^{1}$S. Wang, J. Wu, X. Sun, W. Yuan, W. T. Freeman, J. B. Tenenbaum, and E. H. Adelson are with Computer Science and Artificial Intelligence Laboratory (CSAIL), Massachusetts Institute of Technology, Cambridge, MA 02139, USA}%
\thanks{$^{2}$X. Sun is with Shanghai Jiao Tong University, Shanghai, China}%
\thanks{$^{3}$W. T. Freeman is with Google Research, Cambridge, MA 02142, USA}%
\thanks{$^*$ indicates equal contributions.}
}

\begin{document}

\maketitle
\thispagestyle{empty}
\pagestyle{empty}

\input{text/abstract.tex}

\input{text/intro.tex}

\input{text/related.tex}
\input{text/method.tex}
\input{text/exp.tex}
\input{text/conclusion.tex}

\addtolength{\textheight}{-0cm}   

\noindent\textit{Acknowledgement} This work is supported by ONR MURI N00014-16-1-2007, Toyota, MIT Lincoln Lab, and Facebook.

\bibliographystyle{IEEEtran}
\bibliography{touch_iros}

\end{document}

%% file: packages.tex
\usepackage{color,xcolor}
\usepackage{epsfig}
\usepackage{graphicx}

\usepackage{adjustbox}
\usepackage{array}
\usepackage{booktabs}
\usepackage{colortbl}
\usepackage{float,wrapfig}
\usepackage{hhline}
\usepackage{multirow}
\usepackage{subcaption} 
\usepackage[font={small}]{caption}

\usepackage{amsmath,amsfonts,amssymb}
\usepackage{bm}
\usepackage{nicefrac}
\usepackage{microtype}

\usepackage{changepage}
\usepackage{extramarks}
\usepackage{fancyhdr}
\usepackage{lastpage}
\usepackage{setspace}
\usepackage{soul}
\usepackage{xspace}

\usepackage{url}

\usepackage{algorithm, algorithmic}
\usepackage{enumerate}
\usepackage{todonotes} 
\usepackage{gensymb}

%% file: macros.tex
\newcolumntype{L}[1]{>{\raggedright\let\newline\\\arraybackslash\hspace{0pt}}m{#1}}
\newcolumntype{C}[1]{>{\centering\let\newline\\\arraybackslash\hspace{0pt}}m{#1}}
\newcolumntype{R}[1]{>{\raggedleft\let\newline\\\arraybackslash\hspace{0pt}}m{#1}}

\newcommand{\sect}[1]{Section~\ref{#1}}

\newcommand{\fig}[1]{Figure~\ref{#1}}

\newcommand{\ignorethis}[1]{}

\makeatletter
\DeclareRobustCommand\onedot{\futurelet\@let@token\@onedot}
\def\@onedot{\ifx\@let@token.\else.\null\fi\xspace}

\def\aka{\emph{a.k.a}\onedot}
\def\eg{\emph{e.g}\onedot} 
\def\ie{\emph{i.e}\onedot} 
 
 \def\vs{\emph{vs}\onedot}
 
\def\etal{\emph{et al}\onedot}
\makeatother

\definecolor{MyDarkBlue}{rgb}{0,0.08,1}
\definecolor{MyDarkGreen}{rgb}{0.02,0.6,0.02}
\definecolor{MyDarkRed}{rgb}{0.8,0.02,0.02}
\definecolor{MyDarkOrange}{rgb}{0.40,0.2,0.02}
\definecolor{MyPurple}{RGB}{111,0,255}
\definecolor{MyRed}{rgb}{1.0,0.0,0.0}
\definecolor{MyGold}{rgb}{0.75,0.6,0.12}
\definecolor{MyDarkgray}{rgb}{0.66, 0.66, 0.66}

%% file: text/abstract.tex
\begin{abstract}

Perceiving accurate 3D object shape is important for robots to interact with the physical world. Current research along this direction has been primarily relying on visual observations. Vision, however useful, has inherent limitations due to occlusions and the 2D-3D ambiguities, especially for perception with a monocular camera. In contrast, touch gets precise local shape information, though its efficiency for reconstructing the entire shape could be low. In this paper, we propose a novel paradigm that efficiently perceives accurate 3D object shape by incorporating visual and tactile observations, as well as prior knowledge of common object shapes learned from large-scale shape repositories. We use vision first, applying neural networks with learned shape priors to predict an object's 3D shape from a single-view color image. We then use tactile sensing to refine the shape; the robot actively touches the object regions where the visual prediction has high uncertainty. Our method efficiently builds the 3D shape of common objects from a color image and a small number of tactile explorations (around 10). Our setup is easy to apply and has potentials to help robots better perform grasping or manipulation tasks on real-world objects.

\end{abstract}

%% file: text/intro.tex
\section{Introduction}
\label{sec:intro}

\input{figText/teaser.tex}

For a robot to effectively interact with the physical world, \eg, to recognize, grasp, and manipulate objects, it is highly helpful to know the accurate 3D shape of the objects. 3D shape perception often relies on visual signals; however, using vision alone has fundamental limitations. For example, visual shape perception is often ambiguous due to the difficulties in discriminating the influence of reflection~\cite{Barrow1978Recovering}; real-life occlusions and object self-occlusions also pose challenges to reconstruct full 3D shape from vision. The use of depth sensors alleviates some of these issues, though depth signals can also be too noisy to capture the exact object shape, and depth measurement is largely impacted by the object's color or transparency. 

Touch is another way to perceive 3D shapes. The majority of tactile sensors measure the force distribution or geometry over a small contact area. A robot can use multiple touches, combined with the position and pose of the sensor in each touch, to reconstruct an object's shape without suffering from the ambiguity caused by its surface color or material~\cite{Bohg2017Interactive}. Tactile sensing is however constrained by the size and scale of the sensor: as each touch only gets information of a local region, it may take many touches and a long time to reconstruct the full shape of an object. 

A natural solution is to use tactile sensors to augment vision observations, just as human use fingers---using vision for rough shape reconstruction and touch exploration for shape refinement, especially in occluded regions.  For example, Bjorkman~\etal~\cite{Bjorkman2013Enhancing} explored refining visually perceived shape with touch, where they used a depth camera for a point cloud, a three-finger Schunk Dextrous hand for tactile data, and Gaussian processes for shape prediction.

In this paper, we propose a model that estimates the full 3D shape of common objects from monocular color vision, touch, and learned shape priors. We first use vision to predict the full 3D shape of the object from a monocular color and/or depth image, leveraging the power of 3D deep learning and large-scale 3D shape repositories. Specifically, our model is trained on many 3D CAD models and their RGB-D renderings; it learns to reconstruct a 3D shape from a color image by capturing implicit shape priors throughout the process. It generalizes well to real scenarios, producing plausible 3D shapes from a single image of real-world objects.

We then let the robot touch the object to refine the estimated shape. The tactile sensor we use is a GelSight sensor~\cite{Yuan2017GelSight}, which measures the geometry of local surface with high spatial resolution. By touching object surface with GelSight, the robot obtains additional constraints on the object geometry. Instead of making a local update to the reconstruction for each touch, which is inefficient, we incorporate local tactile constraints to refine the shape globally using the learned shape priors. Moreover, we propose an exploration policy that actively selects the touch point to maximally reduce the uncertainty in the shape prediction. This helps to reduce the number of touches needed.

We aim to make the system efficient and easy to apply. For efficiency, we use only one visual image and a few touch explorations (5--10 touches); for system simplicity, we use a fixed color camera and a tactile sensor on the effector of a 6-DOF robot arm. The setup can be easily applied to other robots as well. 

We test our system on multiple common objects, and show that with a small number of touch exploration, the robot can predict the 3D object shape well. We also present ablation studies to qualitatively and quantitatively validate the effect of our learned shape priors and the active exploration policy. The system can be easily applied to other robots that have a high degree-of-freedom arm and an external color camera. This enables the robot to effectively perceive 3D object shape and to interact with the object.

%% file: figText/teaser.tex
\begin{figure}[t]
    \centering
    \includegraphics[width=\linewidth]{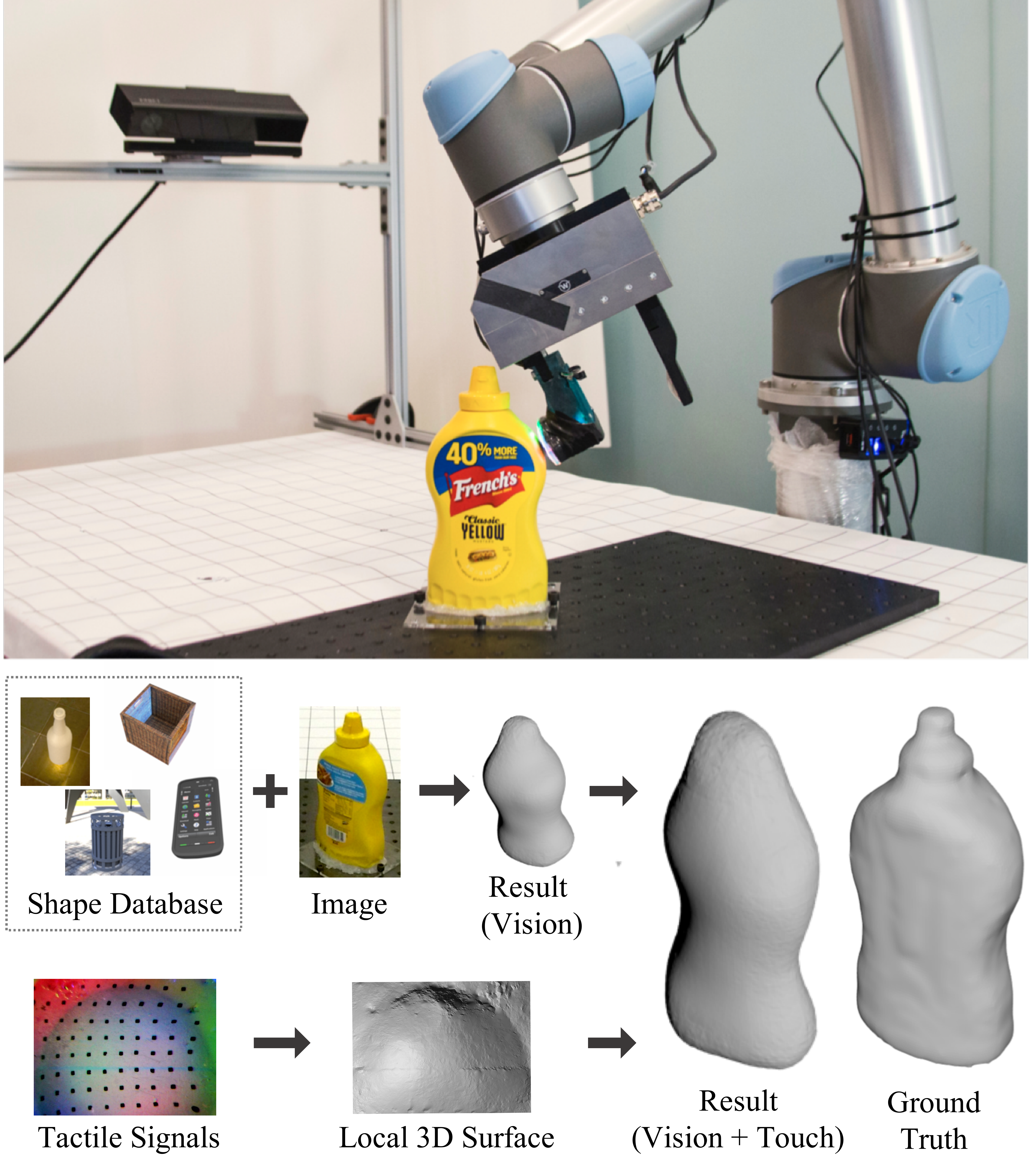}
    \vspace{-20pt}
    \caption{Our model for 3D shape reconstruction. It first reconstructs a rough 3D shape from a single-view color image, leveraging shape priors learned from large-scale 3D shape repositories. It then efficiently incorporates local tactile signals for shape refinement.}
    \vspace{-20pt}
    \label{fig:teaser}
\end{figure}

%% file: text/related.tex
\section{Related Work}
\label{sec:related}

\subsection{3D Shape Reconstruction from Vision}

3D shape completion from depth maps or partial scans has been widely studied in robotics, computer vision, and computer graphics. Traditional methods have attempted to complete shapes with local surface primitives, or to formulate it as an optimization problem, \eg, Poisson surface reconstruction solves an indicator function on a voxel grid via the
Poisson equation~\cite{kazhdan2013screened}. Recently, there have also been a growing number of papers on completing shapes via their structure and regularities~\cite{thrun2005shape} and on learning shape priors in a fully data-driven style~\cite{Firman2016Structured}. In particular, Dai~\etal~\cite{Dai2017Shape} obtained very impressive results on 3D shape completion from partial depth scans by levering 3D convolutional networks and nonparametric patch-based shape synthesis methods.  

The robotics community has explored how shape completion can help grasping using Gaussian Processes~\cite{Dragiev2011Gaussian}. Mahler~\etal~\cite{Mahler2015Gp} explored how Gaussian process implicit surfaces can be used with sequential convex programming for grasp planning. More recently, Varley~\etal~\cite{Varley2017Shape} explored how to better grasp an object by first employing a convolutional neural net for shape completion. 

A more challenging problem is to recover 3D object shape from a single RGB image, without depth information. Solving the problem requires both powerful recognition systems and prior shape knowledge. With large-scale shape repositories like ShapeNet~\cite{Chang2015Shapenet}, researchers have made significant progress on data-driven approaches for shape synthesis, completion, and reconstruction~\cite{Kar2015Category,Wu2016Learning,wueccv,Choy20163d,Soltani2017Synthesizing}.

The problem of 3D reconstruction from RGB data can be reduced into 3D shape completion by first estimating intrinsic images (\eg, depth and surface normal maps) from RGB data~\cite{Barrow1978Recovering}. Some recent papers have studied the problem of depth and surface normal estimation~\cite{Janner2017Self} from a single image. In particular, the vision component of our model builds upon MarrNet~\cite{Wu2017MarrNet}, which jointly estimates intrinsic images and full 3D shape from a color image and has demonstrated good performance on standard benchmarks~\cite{pix3d}.

\subsection{Tactile Sensing for Shape Reconstruction}

In robotics, multi-modal learning has been widely exploited for grasping~\cite{Allen1999Integration}, tracking~\cite{Izatt2017Tracking}, scene layout probing~\cite{Bohg2010Strategies}, and shape recognition with active exploration~\cite{Falco2017Cross}. There has been also research on connecting multi-modality data, \eg, localizing object contact via visual observation~\cite{Luo2015Localizing}, using vision to learn better tactile representations~\cite{Kroemer2011Learning}, and learning the sharing features between vision and tactile~\cite{luo2018vitac}. 

For shape reconstruction in particular, tactile data have also been exploited for both local~\cite{Ottenhaus2016Local}\cite{luo2016iterative} and global shape completion~\cite{Bierbaum2008Robust,pezzementi2011object}, sometimes in a bimanual setting~\cite{Sommer2014Bimanual}. In recent years, researchers started to use active learning for shape reconstruction from tactile sensing~\cite{Driess2017Active,Yi2016Active,Jamali2016Active,martinez2013active}. Luo~\etal recently wrote a comprehensive review article on tactile perception which includes object shape perception~\cite{Luo2017Robotic}. 

Tactile data have been used to complement visual observations for shape reconstruction~\cite{Bjorkman2013Enhancing}, shape reasoning~\cite{Varley2017Visual}, and grasping~\cite{Ilonen2014Three}. Planning has also found its use in shape estimation from visual and tactile data~\cite{Matsubara2017Active}. We refer readers to Bohg~\etal~\cite{Bohg2017Interactive} for a thorough review. These papers, however, directly augment visual observations with tactile signals without leveraging shape priors. In comparison, we use shape priors learned from large-scale shape repositories to efficiently integrate tactile and visual observations.

In this paper, we obtain tactile observations with the GelSight sensor~\cite{Yuan2017GelSight}. GelSight, with its contact surface of a soft elastomer, is able to recover high-fidelity object shape. This makes it particularly useful in object shape reconstruction among tactile sensors. GelSight has also found its in wide applications including physical material modeling~\cite{Yuan2017Connecting} and robot grasping~\cite{Calandra2017Feeling}. 

%% file: text/method.tex
\section{Method}
\label{sec:method}

\input{figText/flowchart.tex}

We reconstruct the 3D shapes of the objects from both vision and touch. The pipeline of the system is described in \fig{fig:flowchart}: we first reconstruct a voxelized rough 3D model of the object from a Kinect color image, and then touch the areas that visual prediction is not of high confidence. The tactile data provide us with the precise location and geometry of the object surface, especially in the occluded areas. These signals can later be posed as constraints to refine the 3D shape. The touch is conducted in a closed-loop exploration process: each time the robot touches the surface location which has the maximum uncertainty in the shape prediction. The policy aims to reduce the times of touches, making the reconstruction more efficient.

\subsection{3D Reconstruction from Vision and Shape Priors}
\label{sec:method_vision}

\input{figText/model.tex}

Our 3D reconstruction model exploits a key intermediate representation---intrinsic images (\aka 2.5D sketches)~\cite{Barrow1978Recovering}. The use of intrinsic images brings in two key advantages. First, it is a unified representation that integrates multi-modal data (RGB images, depth maps, and tactile signals). Using intrinsic images allows us to build a principled framework for multi-model shape reconstruction. Second, color images and 3D shapes become conditionally independent given intrinsic images. When depth data are not available, our formulation decomposes the challenging problem of single-image 3D reconstruction into two simpler ones: intrinsic image estimation and 3D shape completion. This provides us with better reconstruction results from a color image.

Our network, therefore, has two components to recover 3D shape from a color image. The first is a 2.5D sketch estimator (\fig{fig:model}-I), predicting the object's depth, surface normals, and silhouette from the color image; The second is a 3D shape estimator (\fig{fig:model}-II), inferring voxelized 3D object shape from intrinsic images. When depth data is available, we can use them to replace the predicted depth for possible better performance.

\subsubsection{2.5D Sketch Estimation}
\label{sec:step1}

The first component of our network (\fig{fig:model}-I) takes a 2D color image as input and predicts its 2.5D sketches: depth, surface normals, and silhouette. The goal of 2.5D sketch estimation is to distill intrinsic object properties from input images, while discarding properties that are non-essential for the task of 3D reconstruction, such as object texture and lighting.

We use an encoder-decoder network for this step. Our encoder is a ResNet-18~\cite{He2015Deep}, turning a 256$\times$256 RGB image into 384 feature maps, each of size 16$\times$16. Our decoder has three branches for depth, surface normals, and silhouette, respectively. Each branch has four sets of 5$\times$5 transposed convolutional, batch normalization, and ReLU layers, followed by a 1$\times$1 convolutional layer. It outputs at the resolution of 256$\times$256. 

\subsubsection{3D Shape Estimation}
\label{sec:step2}

The second module (\fig{fig:model}-II) infers 3D object shape from estimated 2.5D sketches. Here, the network focuses on learning priors of common shapes. The network architecture is again an encoder and a decoder. It takes a normal image and a depth image as input (both masked by the estimated silhouette), maps them to a 200-dim vector via a modified version of ResNet-18~\cite{He2015Deep}. We changed the average pooling layer into an adaptive average pooling layer, and the output dimension of the last linear layer to 200. The vector then goes through a decoder, consisting of five sets of transposed convolutional, batch normalization, and ReLU layers followed by a transposed convolutional layer and a sigmoid layer to output a 128$\times$128$\times$128 voxel-based reconstruction of the shape. 

\subsection{Tactile Sensing for Shape Refinement}
\label{sec:method_touch}

\input{figText/touch.tex}

Tactile sensing obtains precise information in the local area: the data from the GelSight sensor provide high-resolution 3D geometry of the contact surface, and the position reading from the robot tells the exact location of the touch surface in the global space. The tactile data set solid constraints on the object's shape, and thus help to refine the 3D shape prediction from vision. 

\subsubsection{3D Reconstruction from GelSight}

We can reconstruct the height function $z=f(x,y)$ from the GelSight tactile image~\cite{Yuan2017GelSight}. Under the assumption that the lighting and surface reflectance are evenly distributed, the light intensity $\bm{I}$ at $(x, y)$ can be modeled as 

\begin{equation}
\mathbf{I}(x,y) = \mathbf{R}\left(\frac{\partial f}{\partial x}, \frac{\partial f}{\partial y}\right)
\end{equation}
where $\bm{R}$ is the reflectance function which is a nonlinear function. 

We first build a lookup table to obtain the inverse function $\bm{R}^{-1}$, which maps observed intensity to geometry gradients. A ball with known radius is pressed on the GelSight multiple times to collect data. Then, the gradient can be computed as

\begin{equation}
\left(\frac{\partial f}{\partial x}, \frac{\partial f}{\partial y}\right) = \bm{R}^{-1}(\bm{I}(x, y))
\end{equation}

After calculating the gradients, we reconstruct the height map $z=f(x,y)$ by integrating the gradients. It can be represented as the Poisson equations $(\nabla f)^2 = g$, where 
\begin{equation}
g = \frac{\partial f}{\partial x}\left(\frac{\partial f}{\partial x}\right) + \frac{\partial f}{\partial y}\left(\frac{\partial f}{\partial y}\right).
\end{equation}

We use the fast Poisson solver with the discrete sine transform (DST) to solve it, and get the height-map reconstruction. \fig{fig:touch} shows some examples of the GelSight images and the reconstructed 3D surfaces when contacting different areas on the mustard bottle.

\subsubsection{Registration of World and System Coordinates}

We need to register three coordinate systems: world, robot, and voxel (vision). To align the world the robot frame, we calibrate three points in the real world $\mathbf{x_{w1}}=(0, 0, 0)^T$, $\mathbf{x_{w2}}=(1, 0, 0)^T$, $\mathbf{x_{w3}}=(0, 1, 0)^T$, record their corresponding robot coordinates $\mathbf{x_{r1}}$, $\mathbf{x_{r2}}$, $\mathbf{x_{r3}}$, and calculate the transformation matrix by solving the linear equations
\begin{equation}
\mathbf{X_r} = \mathbf{R_{r}}\cdot \mathbf{X_w} + \mathbf{T_{r}},
\end{equation}
where $\mathbf{X_r=[x_{r1}, x_{r2}, x_{r3}]}$, $\mathbf{X_w=[x_{w1}, x_{w2}, x_{w3}]}$, $\mathbf{R_{r}}$ is the rotation matrix, and $\mathbf{T_{r}}$ is the translation vector. 

To align voxels with the world frame, we use the correspondence of a fixed point $\mathbf{o}$, axes $\mathbf{a_x}, \mathbf{a_y}, \mathbf{a_z}$, and the scale $s$ to calculate the transformation. The bottom center of the voxels $\mathbf{o_v}$ is aligned with the fixed point on the table in the world frame $\mathbf{o_w}$. The axes can be calculated based on the camera's position and orientation. In our setting, $\mathbf{a_x} = (-1, 0, 0)^T$, $\mathbf{a_y} = (0, 1, 0)^T$, $\mathbf{a_z}=(0, 0, 1)^T$. The scale of each voxel can be calculated by $s = n_p\times l_p$, where $n_p$ is the number of corresponding pixels to each voxel and $l_p$ is the length of each pixel in the real world. Then the rotation matrix $\mathbf{R_v}=s\cdot \begin{bmatrix}\mathbf{a_x}, \mathbf{a_y},\mathbf{a_z}\end{bmatrix}^T$. The transformation between world and voxel coordinate can be represented as
\begin{equation}
    \mathbf{x_w} = \mathbf{R_v} \cdot (\mathbf{x_v} - \mathbf{o_v}) + \mathbf{o_w}.
\end{equation}

After registration, we can map touches into corresponding voxels and control the robot arm to touch the target regions in the real world.

\subsubsection{Updating Shape Reconstruction with Touch}

We then present how we update the model's prediction with tactile signals, after converting them into surface normals, and registering them into the system coordinates. The key observation here is to design a differentiable loss function that enables fine-tuning with back-propagation.

\fig{fig:loss} illustrates our design. Given a 3D point in space and its normal vector $\bm{n}$, we gradually move the robot arm toward the destination, unless it touches a solid object halfway between. Either way, we obtain signals on whether the 3D voxels along the trajectory are occupied. We use $v_{\bm{p}}$ to represent the value at position $\bm{p}$ in a 3D voxel grid, where $v_{\bm{p}}\in[0,1]$. Assume the GelSight sensor suggests the voxel $\bm{p}_0=\{x_0,y_0,z_0\}$ is filled (\fig{fig:loss}b). Our differentiable loss tries to encourage the voxel's value to be 1, and all voxels in front of it, along the direction $\bm{n}$, to be $0$. This ensures the estimated 3D shape matches the obtained tactile signals. The differentiable loss for a voxel $\bm{p}$ is defined as
\begin{equation}
        L(v_{\bm{p}}) = 
        \begin{cases}
            v^2_{\bm{p}}, & \bm{p} = \bm{p}_0 + k\bm{n}, \quad \forall k<0 \\
            (1 - v_{\bm{p}})^2, & \bm{p} = \bm{p}_0 \\
            0, & \text{otherwise} \\
        \end{cases}.
\end{equation}
The gradients are
\begin{equation}
        \frac{\partial{L(v_{\bm{p}})}}{\partial{v_{\bm{p}}}} =
        \begin{cases}
            2 v_{\bm{p}}, & \bm{p} = \bm{p}_0 + k\bm{n}, \quad \forall k<0 \\
            2 (v_{\bm{p}} - 1), & \bm{p} = \bm{p}_0 \\
            0, & \text{otherwise} \\
        \end{cases}.
\end{equation}
The loss and gradients can be similarly derived when the GelSight sensor suggests the voxel $\bm{p}_0$ is empty (\fig{fig:loss}a).

After collecting touch signals, we compute losses and back-propagate gradients to the latent vector from the 2.5D sketch encoder. We then update it (with a learning rate of 0.001) and use the shape decoder to get a new shape. We repeat this process for 10 iterations for each touch.

\subsection{Policy for Active Tactile Exploration}
\label{sec:method_policy}

\input{figText/loss.tex}

We here describe our policy that automatically discovers the most uncertain region of the prediction for the next tactile exploration. Since the value of each voxel $v_{i,j,k}$ is the output of the sigmoid function which indicates the existing probability, the network's confidence score of voxel $v_{i,j,k}$ is defined as $c_{i,j,k}=|v_{i,j,k}-0.5|$. We therefore would like to find a region $S$ that is of the same size as the GelSight sensor and minimizes $\sum_{(i,j,k)\in S}c_{i,j,k}$.
sear
This seemingly simple problem is challenging as the region $S$ can be of any orientation, and we want the optimization to be fast. Our algorithm is based on integral maps. Given a plane, we sample a 2D grid on the plane and assign each point's confidence score $f_{p,q}$ as its closest voxel's confidence score, as shown in \fig{fig:policy}a. We then compute the integral maps on the 2D grid; specifically, we have
$g_{p,q} = \sum_{i=1}^p\sum_{j=1}^q f_{i,j}$.
As 
\begin{equation}
g_{p,q}=f_{p,q}+g_{p-1,q}+g_{p,q-1}-g_{p-1,q-1},    
\end{equation}
we can compute the matrix $\bm{G}$ in $O(N^2)$ time, where $N$ is the length of the voxel grid. 

As the size of the GelSight sensor $S=k\times k$ is known, we can then find the region $S$ with a minimal summed confidence score using $\bm{G}$, again in $O(N^2)$. This is because for a particular region $[p+1, p+k]\times[q+1,q+k]$, as shown in \fig{fig:policy}b, we can compute its regional sum in $O(1)$ as
\begin{equation}
\sum_{i=1}^k\sum_{j=1}^k f_{p+i,q+j}=g_{p+k,q+k}-g_{p,q+k}-g_{p+k,q}+g_{p,q}.
\end{equation}
Finally, we in parallel evaluate multiple planes by searching over yaws (every $90^{\circ}$) and pitches (every $10^{\circ}$). 

\input{figText/policy.tex}

%% file: figText/flowchart.tex
\begin{figure}[t]
    \centering
    \includegraphics[width=\linewidth]{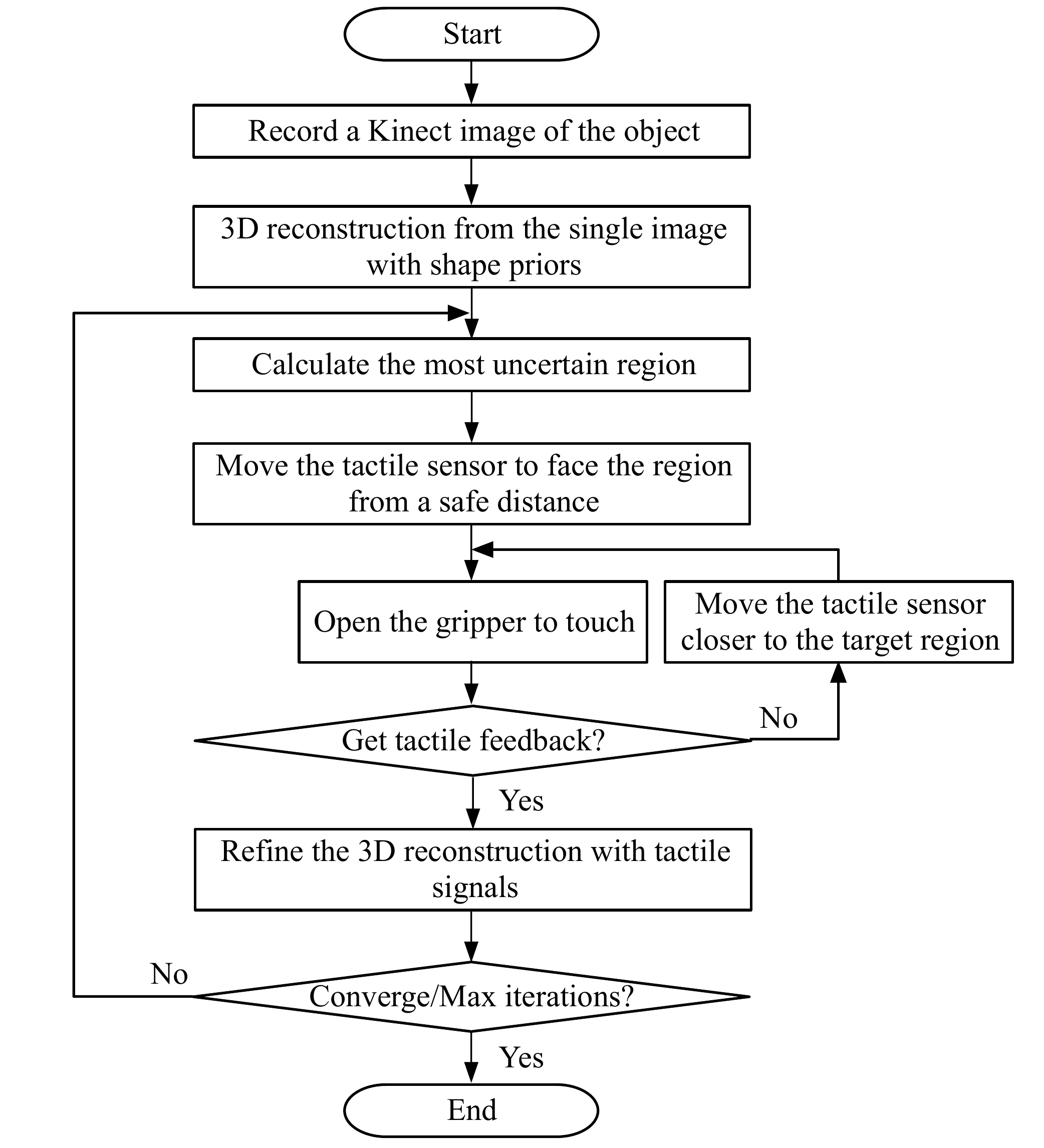}
    \caption{An overview of our interactive system that estimates 3D shape from monocular vision, touch, and shape priors.}
    \vspace{-15pt}
    \label{fig:flowchart}
\end{figure}

%% file: figText/model.tex
\begin{figure*}[t]
    \centering
    \includegraphics[width=\linewidth]{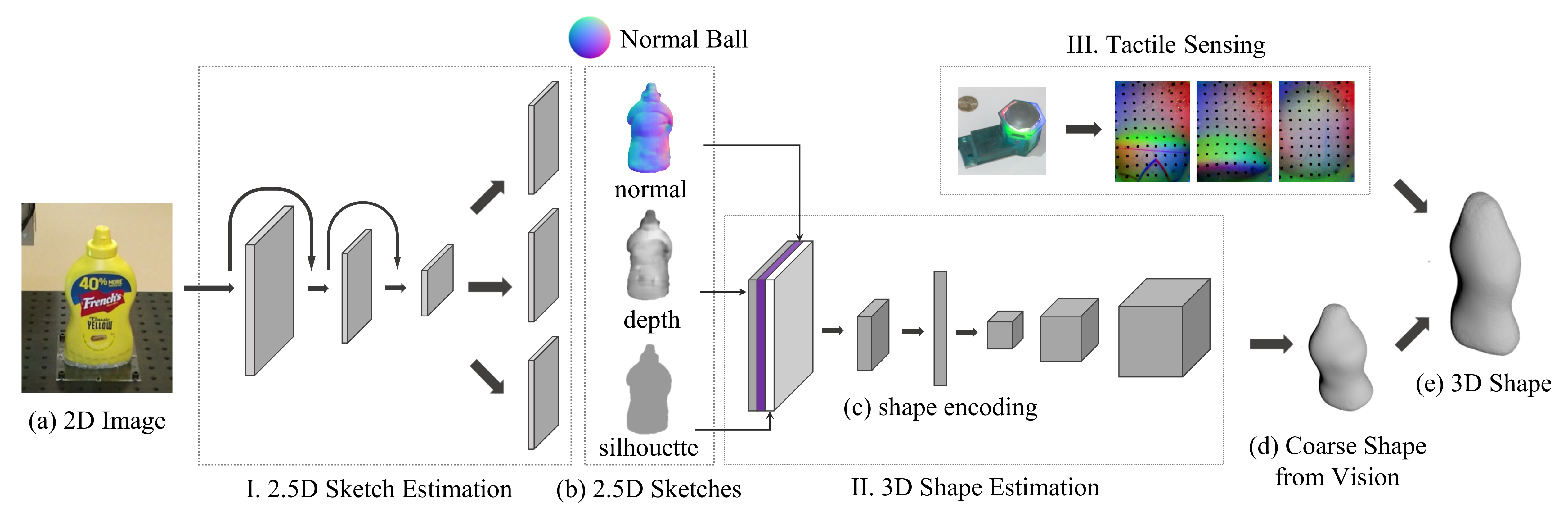}
    \vspace{-15pt}
    \caption{Our model has three major components. It first estimates the object's 2.5D sketches (depth, surface normals, and silhouette) from a single RGB image. It then recovers a rough 3D shape from them. Third, it integrates tactile signals to update the latent shape encoding and to generate a refined 3D shape.}
    \vspace{-15pt}
    \label{fig:model}
\end{figure*}

%% file: figText/touch.tex
 \begin{figure}[t]
    \centering
    \includegraphics[width=\linewidth]{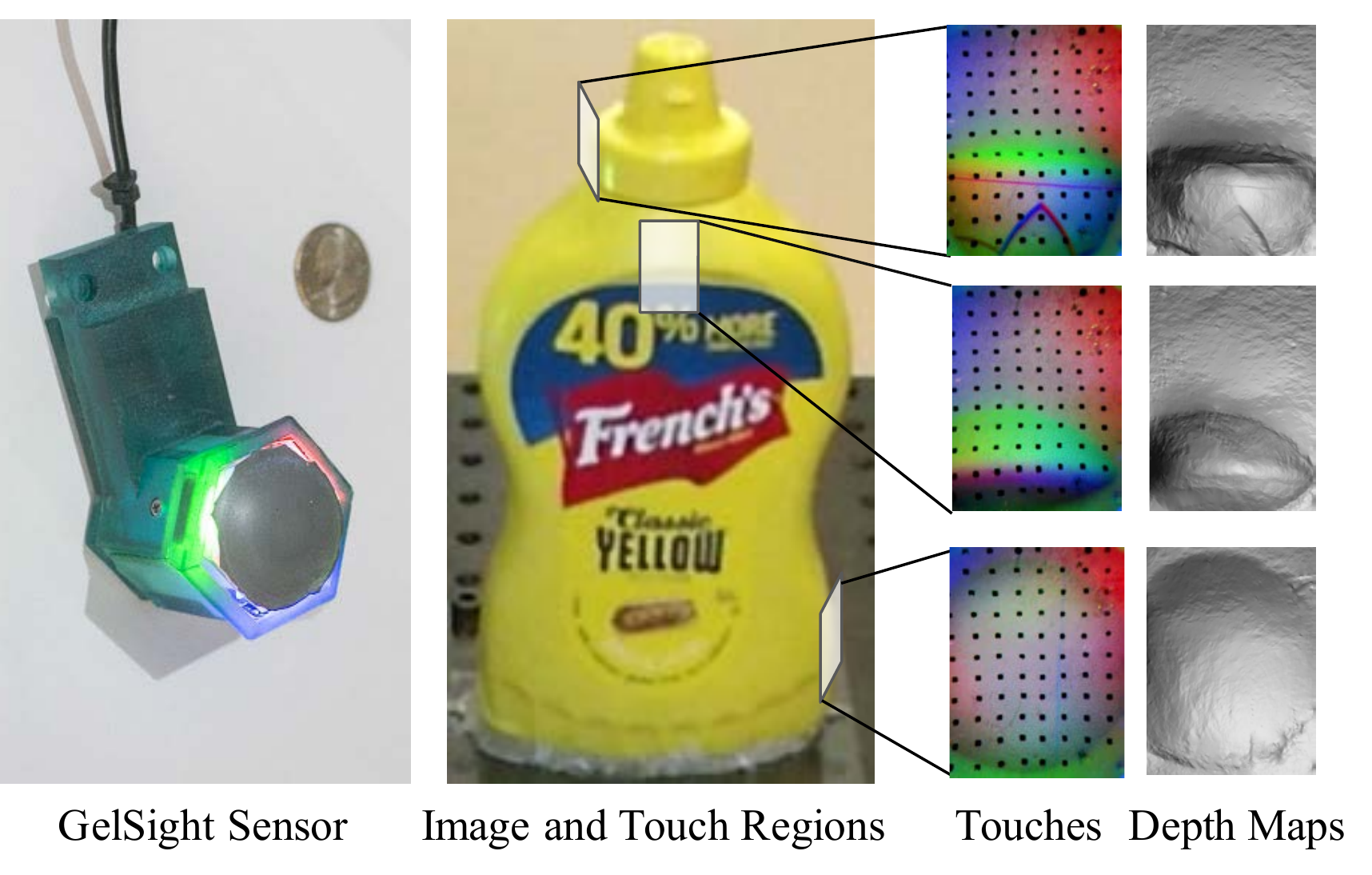}
    \vspace{-15pt}
    \caption{Tactile signals on different parts of the object and the corresponding 3D reconstructions}
    \vspace{-20pt}
    \label{fig:touch}
\end{figure}

%% file: figText/loss.tex
\begin{figure}[t]
    \centering
    \includegraphics[width=\linewidth]{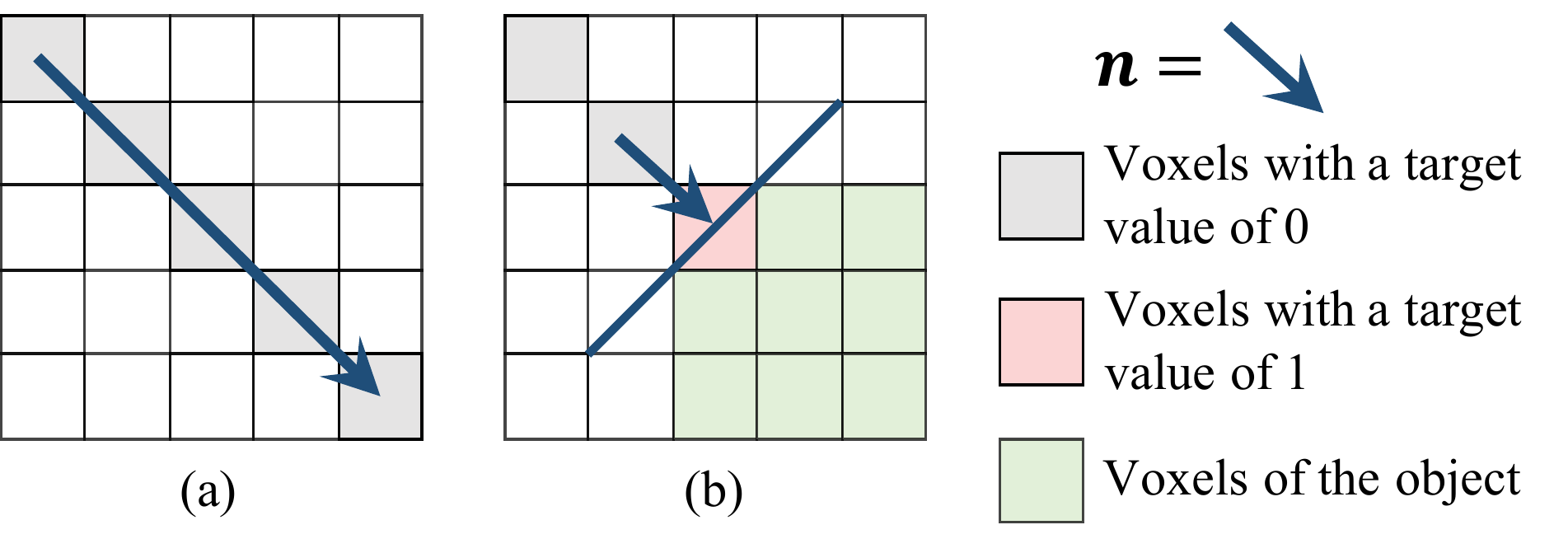}
    \vspace{-15pt}
    \caption{Reprojection loss for touches. (a) When the sensor makes a touch attempt but fails to reach the object, the voxels along its trajectory should all be 0. (b) When the sensor contacts the object, the corresponding voxels should be 1, and all voxels in front of it along the trajectory should be 0.}
    \vspace{-10pt}
    \label{fig:loss}
\end{figure}

%% file: figText/policy.tex
\begin{figure}[t]
    \centering
    \includegraphics[width=\linewidth]{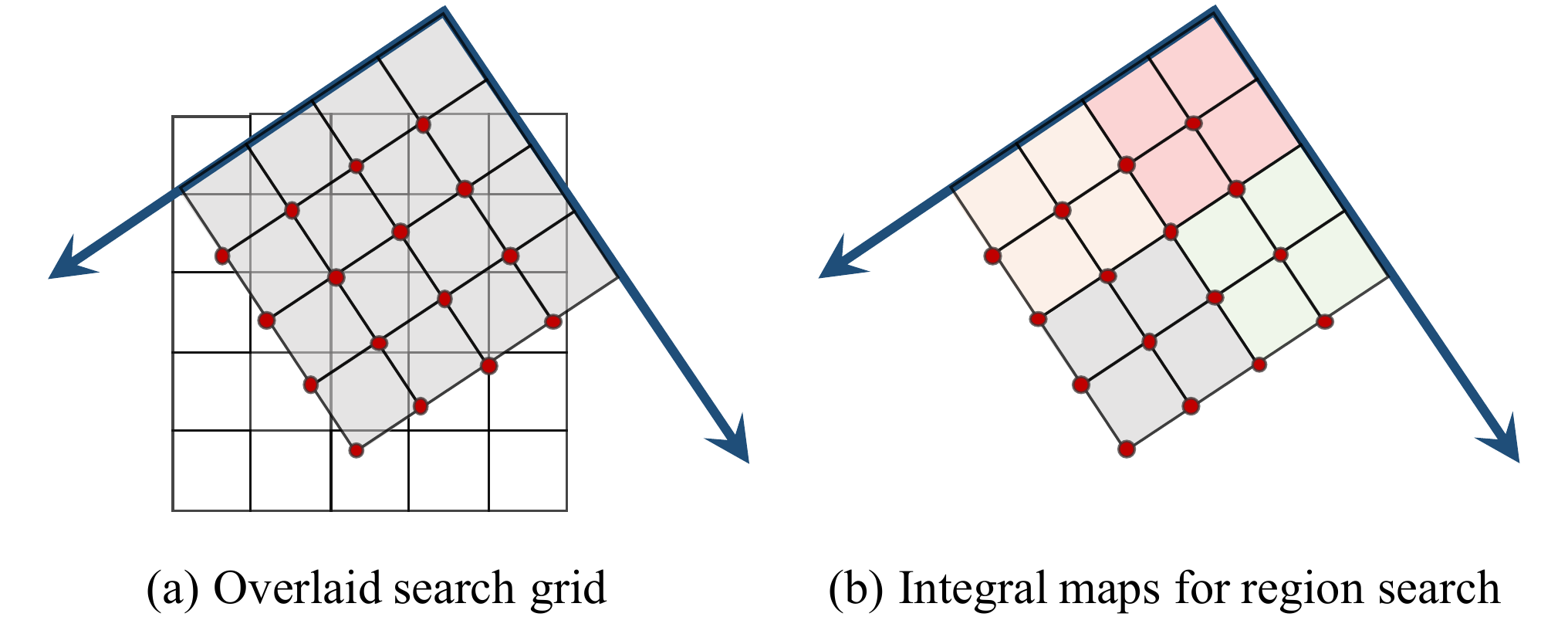}
    \vspace{-15pt}
    \caption{Our policy on finding the next place to touch. (a) A 2D search grid overlaid on the voxel grid, where the confidence values of the voxel prediction are assigned to the search grid. (b) After the assignment, we compute the integral map and use it to efficiently search for the region of maximal uncertainty. See text for details.}
    \vspace{-15pt}
    \label{fig:policy}
\end{figure}

%% file: text/exp.tex
\section{Experiments}
\label{sec:exp}

We now present experimental results. We first introduce our robot platform setup and how we generate training data for the networks. We then discuss our main results---how we reconstruct high-quality 3D shapes with vision, touch, and shape priors. Further, we conduct ablation studies to understand the contributions of each model component: how shape priors and the active exploration policy help to reconstruct shapes more efficiently, and how well our system adapts to RGB and depth data.

\input{figText/results.tex}

\subsection{Robotic System Setup}

The robotic system includes a 6-DOF robot arm, a GelSight tactile sensor, and a Kinect 2 (as shown in  \fig{fig:teaser}). The GelSight sensor is mounted on a WSG 50 parallel gripper for the convenience. The target object is fixed to an optical breadboard in the robot's working space so that it will keep static during the interaction with the robot.

The robot arm is a UR5 from Universal Robotics with a reach radius of 850mm. The WSG 50 gripper is a parallel gripper from Weiss Robotics with force feedback. We do not use the gripper for gripping the objects, but we use the gripper's force feedback to alert collision of the sensor so that we install the GelSight sensor outwards in order to better touch the objects. The GelSight sensor we apply is the version introduced in \cite{dong2017improved}. It captures the surface geometry of a contact area of 19mm$\times$14mm with a resolution of 640$\times$480 and a frequency of 30Hz. The raw output from the sensor is in the format of an image, and we reconstruct the 2.5D topography of the surface from it. The Kinect 2 captures RGB-D images of the target area, and is fixed on the side of the table at a 45.72cm height and a $30^{\circ}$ tilt angle.

\subsection{Conducting Touch without Collision}

When touching the object surface, the robot should carefully avoid collision with the object. This is especially the case in our setup, as the initial 3D reconstruction can be imprecise, and the robot does not have much effective contact feedback other than the sensing surface of the GelSight sensor. We make the robot progressively head toward the target region from distance in the direction of the surface normal. In each touch attempt, the approach is conducted by the slow opening of the parallel gripper, so that the force feedback from the gripper's current provides a protection of the collision, especially when the collision does not happen on the GelSight's sensing area. At the same time, we also plan the motion of the robot when transferring between different touch attempt to avoid interfering with the object. Our basic strategy is to take a detour in the high-up area when changing the target positions. But we also calculate the radial angles between the two target locations. When the angle is small, it indicates that the two locations are close, and it is safe for the robot to move directly to the second location to save time.

\subsection{Dataset}

We generate synthetic training data of paired images and 3D shapes for networks to learn shape priors. We use Mitsuba~\cite{Jakob2010Mitsuba} to render fourteen object categories (bag, bottle, bowl, camera, can, cap, computer keyboard, earphone, helmet, jar, knife, laptop, mug, remote control) in ShapeNet~\cite{Chang2015Shapenet} from 20 random views using three types of backgrounds: 1/3 on a clean, white background, 1/3 on high-dynamic-range backgrounds with illumination channels, and 1/3 on backgrounds randomly sampled from the SUN database~\cite{Xiao2010Sun}. For each object in each view, we render an RGB image and its depth, surface normal, and silhouette. We augment our training data by color and light jittering during training. 

We train the 2.5D sketch estimator and the 3D shape estimator separately on synthetic images. The 2.5D sketch estimator is trained using the ground truth surface normal, depth, and silhouette images with an L2 loss. The 3D shape estimator is trained using ground truth voxels and a binary cross-entropy loss. We implement our model in PyTorch. We use the Adam optimizer~\cite{Kingma2015Adam} with $\beta_1=0.5$, $\beta_2=0.9$ and a learning rate of $5\times10^{-4}$ for the 2.5D sketch estimator, and stochastic gradient descent with a learning rate of $2\times10^{-2}$ and a momentum of $0.9$ for the 3D shape estimator. For visualization, bilateral filters are applied to remove aliasing~\cite{jones2003non}. 

\subsection{Results}

We show the main results in \fig{fig:results}. From a single RGB image, our learned model correctly segments the object and produces a rough 3D shape estimation. We then let the robot automatically touch the objects and use the tactile signals to further refine the shape. For the sugar box in row 3, we use a prior learned on box-like shapes instead of all fourteen categories. An ablation study is presented in \sect{sec:ablation}.

Our system works well on a variety of object shapes. Each example shown in the figure has its distinct shape, and our model works well on all of them. For example, our model recovers the fine curvature of the spray bottles. As our model does not require a depth image as input, it can deal with transparent objects like the water bottle (though it can still use Kinect depth when available, as shown in \sect{sec:rgbd}).  

\input{figText/rgbd.tex}
\input{figText/ablation.tex}

\subsection{Shape Priors and the Exploration Policy}
\label{sec:ablation}

We then present three ablation studies to understand how the learned priors and the active exploration policy contribute to its final performance. First, we compare our model with two variants: Direct Edit and Random Policy. The first one does not use shape priors; instead, it directly uses the tactile signals to edit the voxelized shape, \ie changing the values of the touched voxels to 1 and the voxels in front of them to 0. The second does not use our policy. It randomly chooses where to touch within the object's bounding box. The performance of the second baseline has large variance due to its randomness. For quantitative evaluation, we run it 10 times and compute the mean of its scores.

\fig{fig:ablation} shows qualitative results. Both the policy and the shape priors help to obtain an accurate shape estimation much faster, significantly reducing the number of touches required. Without the priors, each touch can only be used to update a local region of the shape; without the policy, the shape may become significantly worse before eventually getting better.

We further quantitatively compare the shape obtained after each update with the ground truth shape. Our metric is the classic Chamfer distance (CD)~\cite{Barrow1977Parametric}, widely used in the computer graphics community for measuring shape similarity. For each point in each cloud, CD finds the nearest point in the other point set, and sums the distances up. 

We show quantitative results in \fig{fig:curves}. Here, we also have a human policy, where humans select the position of the next touch. This can be seen as an upper bound of possible performance. Our full model achieves a low Chamfer distance after a few touches, close to the human, while the baselines (w/o policy or priors) take much longer. 

\input{figText/prior.tex}

We also evaluate how priors learned on different training sets affect results. \fig{fig:prior} shows that for the sugar box, a network trained on general shapes predicts a less accurate shape, which is later corrected by touches; in contrast, a network trained on box-like shapes gives better results. This reveals an interesting future direction: it will be helpful to classify the object's type from vision, which may inform the most efficient policy and prior.

\input{figText/curves.tex}

\subsection{RGB vs RGB-D Input}
\label{sec:rgbd}

We finally evaluate how our model works on RGB \vs RGB-D data, to better understand its practical applicability. \fig{fig:rgbd} reveals that our method can use either our estimated depth maps or Kinect depth maps. A Kinect depth map can be helpful if it is accurate: for example, the initial reconstruction of the left bottle is flatter (and therefore better) using the Kinect depth map. However, Kinect depth maps can also be unreliable: it fails to estimate the depth of the transparent water bottle. If we purely rely on Kinect depth, our reconstruction would not be as accurate as our current formulation, which is able to recover 3D shape purely from a color image and touch.

%% file: figText/results.tex
\begin{figure*}[t]
    \centering
    \includegraphics[width=\linewidth]{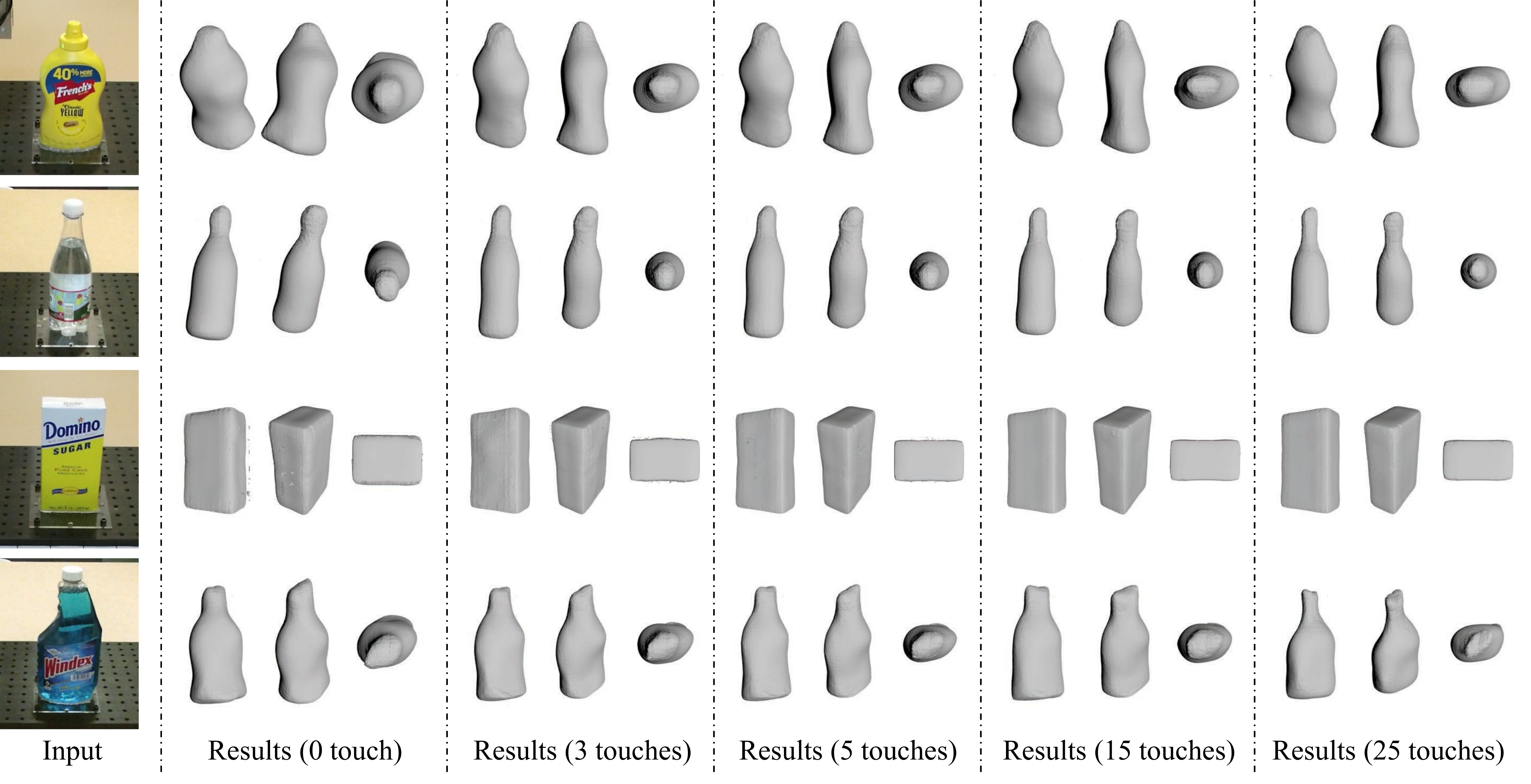}
    \vspace{-15pt}
    \caption{Results on 3D shape perception. From a single RGB image, our model recovers a rough 3D shape using shape priors. The reconstruction often captures the basic geometry, but deviates from the actual shape in various ways. The results improve gradually with touch signals. For example, for the bell-shaped bottle in the last row, the initial reconstruction is too fat (best seen from the top-down view). With tactile signals, our model recovers its flat shape. Our system also corrects object pose, as shown in the water bottle case.}
    \vspace{-18pt}
    \label{fig:results}
\end{figure*}

%% file: figText/rgbd.tex
\begin{figure*}[t]
    \centering
    \includegraphics[width=\linewidth]{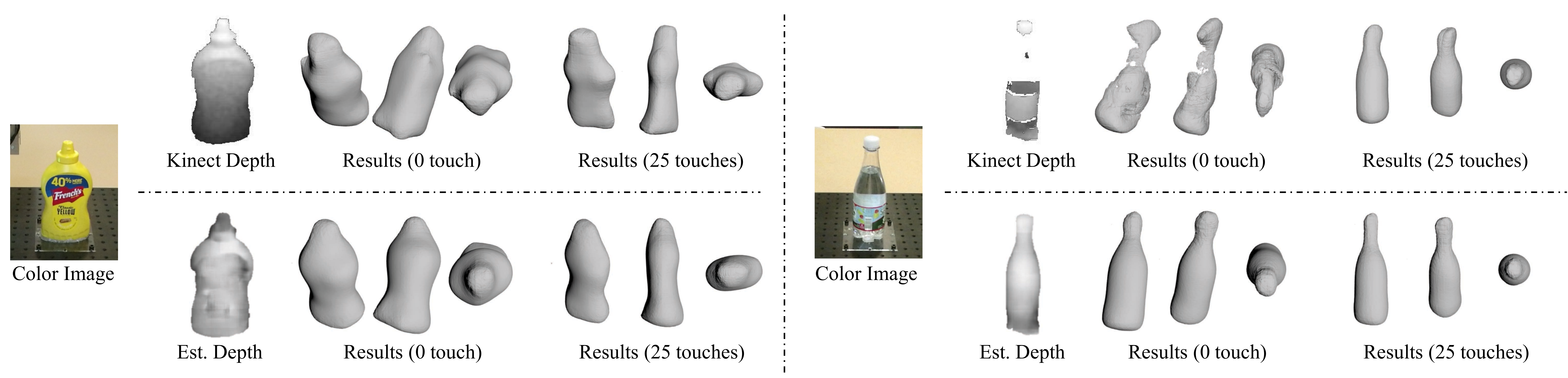}
    \vspace{-15pt}
    \caption{Our method can use either our estimated depth maps or Kinect depth maps. A Kinect depth map can be helpful if it is accurate: for example, the initial reconstruction of the left bottle is flatter using the Kinect depth map. However, if we purely rely on Kinect depth, our reconstruction would not be as accurate when the Kinect depth is inaccurate (see the transparent water bottle).} 
    \vspace{-18pt}
    \label{fig:rgbd}
\end{figure*}

%% file: figText/ablation.tex
\begin{figure}[t]
    \centering
    \includegraphics[width=\linewidth]{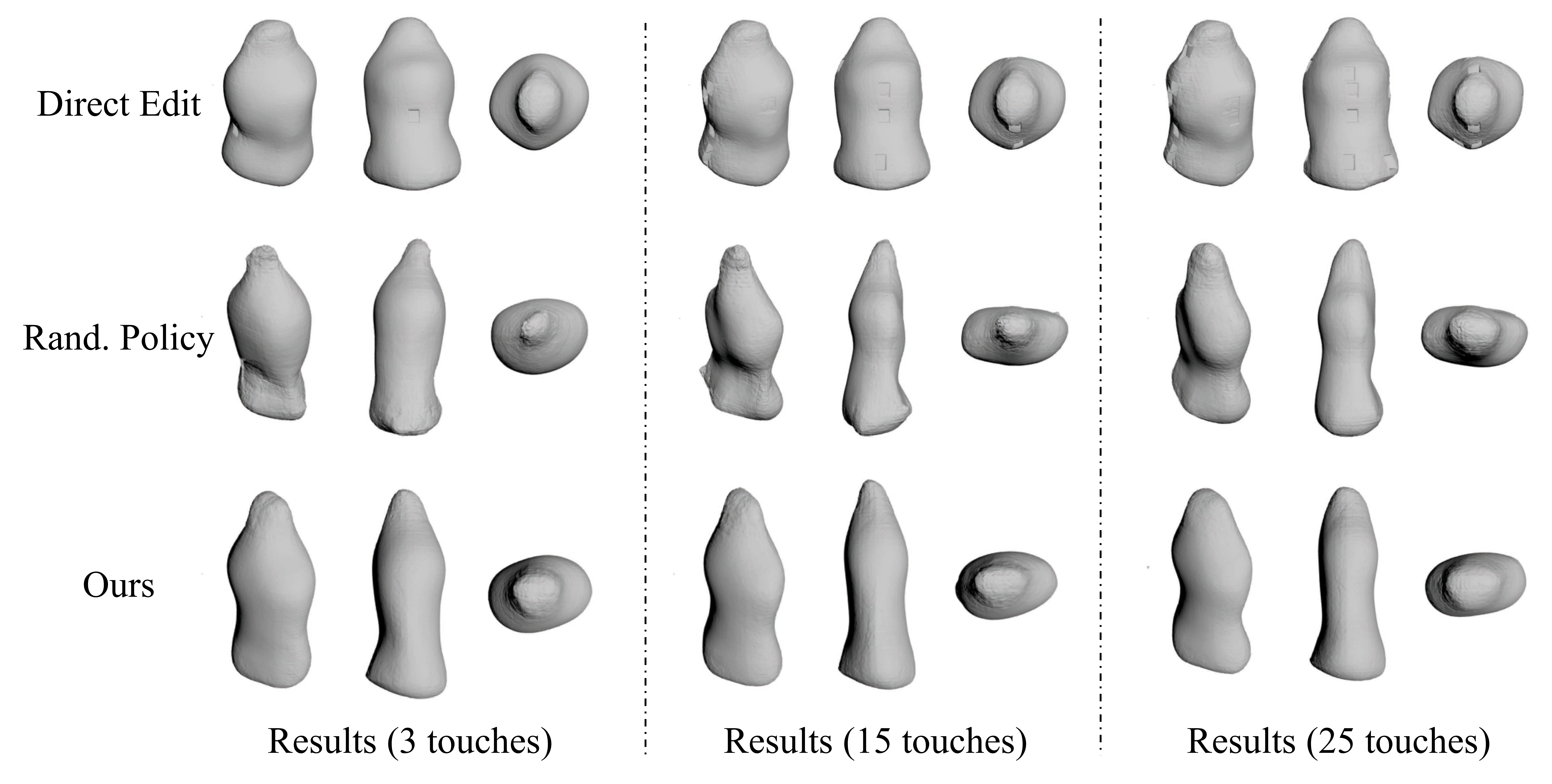}
    \vspace{-15pt}
    \caption{We show the effects of shape priors and the policy. If we Direct Edit the voxels' value (not using learned priors to update), each touch can only be used to update the shape locally. The shape does not change much even after many touches. With Random Policy, it takes longer for the model to obtain fine shape structure.}
    \vspace{-12pt}
    \label{fig:ablation}
\end{figure}

%% file: figText/prior.tex
\begin{figure}[t]
    \centering
    \includegraphics[width=0.95\linewidth]{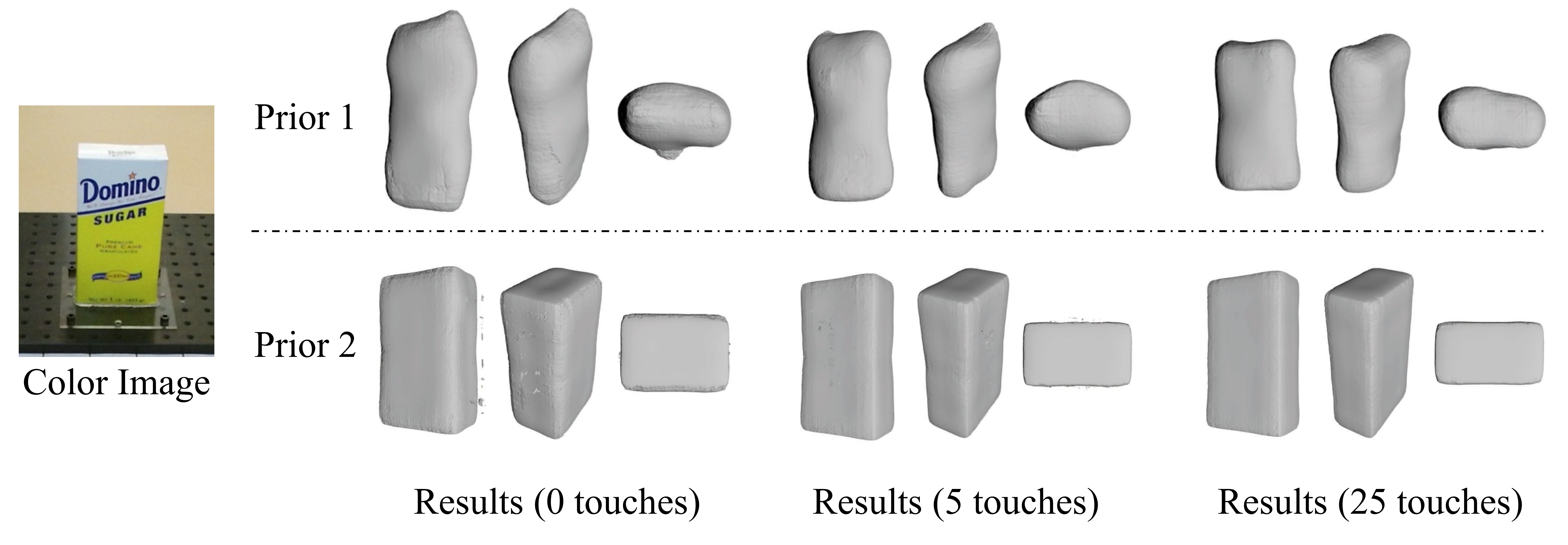}
    \vspace{-6pt}
    \caption{The two priors on the sugar box. A network trained on general shapes predicts a less accurate shape, which is later corrected by touches. A network trained on box-like shapes gives better results.}
    \vspace{-20pt}
    \label{fig:prior}
\end{figure}

%% file: figText/curves.tex
\begin{figure}[t]
    \centering
    \includegraphics[width=0.85\linewidth]{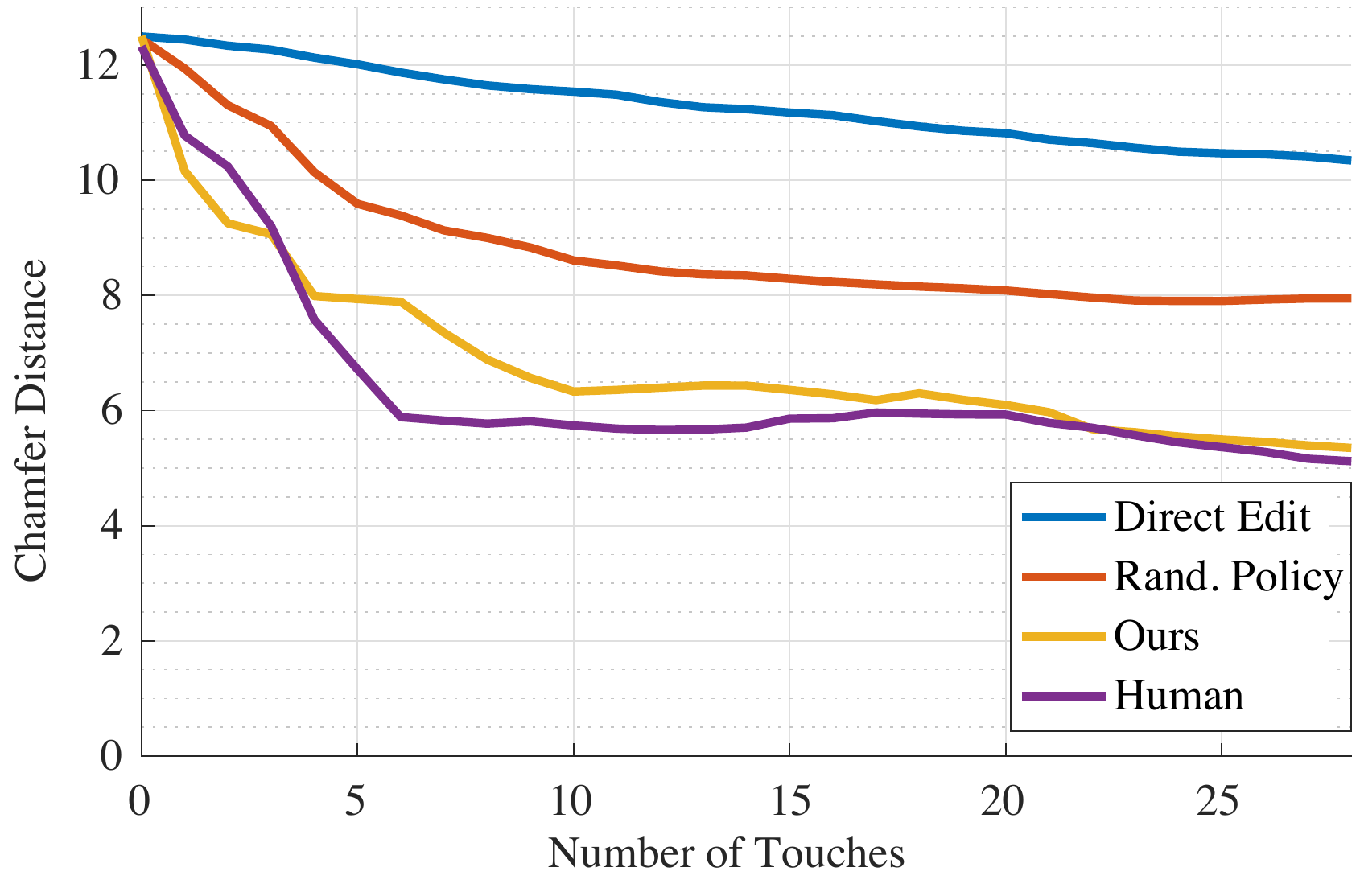}
    \vspace{-5pt}
    \caption{Shape estimation accuracy with respect to the number of touches, measured in Chamfer distance. Our policy recovers the shape accurately and efficiently. With Random Policy, it takes much longer to reconstruct a reasonable shape; if we Direct Edit the voxels' value (not using learned priors to update), the object is hardly updated after each touch. The Human method asks a human to manually select where to touch for each step and can be seen as an upper bound of an automatic algorithm's performance.}
    \vspace{-20pt}
    \label{fig:curves}
\end{figure}

%% file: text/conclusion.tex
\vspace{-3pt}
\section{Conclusion} 
\label{sec:con}
    
We have presented a novel model for 3D shape perception that integrates visual and tactile signals with learned shape priors. Our model uses intrinsic images as the intermediate representation to unify multi-modal signals. We have also proposed an active exploration policy to search for the most informative touches. Our model performs well on real objects, recovering their 3D shape accurately. Ablation studies verify that the use of touch priors and the exploration policy enables more efficient shape recovery. Our model works well with RGB and RGB-D data, and can handle transparent objects.

We hope our approach can inspire future research in fusing common sense knowledge into building object models: the idea of learning an object prior can be extended to not only model shapes, but objects' physical attributes; we can also refine the learned object prior through interaction~\cite{Bohg2017Interactive}. 

%% file: touch_iros.bbl
\begin{thebibliography}{10}
\providecommand{\url}[1]{#1}
\csname url@samestyle\endcsname
\providecommand{\newblock}{\relax}
\providecommand{\bibinfo}[2]{#2}
\providecommand{\BIBentrySTDinterwordspacing}{\spaceskip=0pt\relax}
\providecommand{\BIBentryALTinterwordstretchfactor}{4}
\providecommand{\BIBentryALTinterwordspacing}{\spaceskip=\fontdimen2\font plus
\BIBentryALTinterwordstretchfactor\fontdimen3\font minus
  \fontdimen4\font\relax}
\providecommand{\BIBforeignlanguage}[2]{{%
\expandafter\ifx\csname l@#1\endcsname\relax
\typeout{** WARNING: IEEEtran.bst: No hyphenation pattern has been}%
\typeout{** loaded for the language `#1'. Using the pattern for}%
\typeout{** the default language instead.}%
\else
\language=\csname l@#1\endcsname
\fi
#2}}
\providecommand{\BIBdecl}{\relax}
\BIBdecl

\bibitem{Barrow1978Recovering}
H.~G. Barrow and J.~M. Tenenbaum, ``Recovering intrinsic scene characteristics
  from images,'' \emph{Computer Vision Systems}, 1978.

\bibitem{Bohg2017Interactive}
J.~Bohg, K.~Hausman, B.~Sankaran, O.~Brock, D.~Kragic, S.~Schaal, and G.~S.
  Sukhatme, ``Interactive perception: Leveraging action in perception and
  perception in action,'' \emph{IEEE Trans. Robotics}, vol.~33, no.~6, pp.
  1273--1291, 2017.

\bibitem{Bjorkman2013Enhancing}
M.~Bjorkman, Y.~Bekiroglu, V.~Hogman, and D.~Kragic, ``Enhancing visual
  perception of shape through tactile glances,'' in \emph{IEEE International
  Conference on Intelligent Robots and Systems (IROS)}, 2013.

\bibitem{Yuan2017GelSight}
W.~Yuan, S.~Dong, and E.~H. Adelson, ``Gelsight: High-resolution robot tactile
  sensors for estimating geometry and force,'' \emph{Sensors}, vol.~17, no.~12,
  p. 2762, 2017.

\bibitem{kazhdan2013screened}
M.~Kazhdan and H.~Hoppe, ``Screened poisson surface reconstruction,'' \emph{ACM
  Trans. Graph.}, vol.~32, no.~3, p.~29, 2013.

\bibitem{thrun2005shape}
S.~Thrun and B.~Wegbreit, ``Shape from symmetry,'' in \emph{IEEE International
  Conference on Computer Vision (ICCV)}, 2005.

\bibitem{Firman2016Structured}
M.~Firman, O.~Mac~Aodha, S.~Julier, and G.~J. Brostow, ``Structured prediction
  of unobserved voxels from a single depth image,'' in \emph{IEEE Conference on
  Computer Vision and Pattern Recognition (CVPR)}, 2016.

\bibitem{Dai2017Shape}
A.~Dai, C.~R. Qi, and M.~Nie{\ss}ner, ``Shape completion using
  3d-encoder-predictor cnns and shape synthesis,'' in \emph{IEEE Conference on
  Computer Vision and Pattern Recognition (CVPR)}, 2017.

\bibitem{Dragiev2011Gaussian}
S.~Dragiev, M.~Toussaint, and M.~Gienger, ``Gaussian process implicit surfaces
  for shape estimation and grasping,'' in \emph{IEEE International Conference
  on Robotics and Automation (ICRA)}, 2011.

\bibitem{Mahler2015Gp}
J.~Mahler, S.~Patil, B.~Kehoe, J.~Van Den~Berg, M.~Ciocarlie, P.~Abbeel, and
  K.~Goldberg, ``Gp-gpis-opt: Grasp planning with shape uncertainty using
  gaussian process implicit surfaces and sequential convex programming,'' in
  \emph{IEEE International Conference on Robotics and Automation (ICRA)}, 2015.

\bibitem{Varley2017Shape}
J.~Varley, C.~DeChant, A.~Richardson, A.~Nair, J.~Ruales, and P.~Allen, ``Shape
  completion enabled robotic grasping,'' in \emph{IEEE International Conference
  on Intelligent Robots and Systems (IROS)}, 2017.

\bibitem{Chang2015Shapenet}
A.~X. Chang \emph{et~al.}, ``{Shapenet: An information-rich 3d model
  repository},'' \emph{arXiv:1512.03012}, 2015.

\bibitem{Kar2015Category}
A.~Kar, S.~Tulsiani, J.~Carreira, and J.~Malik, ``Category-specific object
  reconstruction from a single image,'' in \emph{IEEE Conference on Computer
  Vision and Pattern Recognition (CVPR)}, 2015.

\bibitem{Wu2016Learning}
J.~Wu, C.~Zhang, T.~Xue, W.~T. Freeman, and J.~B. Tenenbaum, ``{Learning a
  Probabilistic Latent Space of Object Shapes via 3D Generative-Adversarial
  Modeling},'' in \emph{Neural Information Processing Systems (NIPS)}, 2016.

\bibitem{wueccv}
J.~Wu, C.~Zhang, X.~Zhang, Z.~Zhang, W.~T. Freeman, and J.~B. Tenenbaum,
  ``Learning shape priors for 3d shape completion and reconstruction,'' in
  \emph{European Conference on Computer Vision (ECCV)}, 2018.

\bibitem{Choy20163d}
C.~B. Choy, D.~Xu, J.~Gwak, K.~Chen, and S.~Savarese, ``3d-r2n2: A unified
  approach for single and multi-view 3d object reconstruction,'' in
  \emph{European Conference on Computer Vision (ECCV)}, 2016.

\bibitem{Soltani2017Synthesizing}
A.~A. Soltani, H.~Huang, J.~Wu, T.~D. Kulkarni, and J.~B. Tenenbaum,
  ``Synthesizing 3d shapes via modeling multi-view depth maps and silhouettes
  with deep generative networks,'' in \emph{IEEE Conference on Computer Vision
  and Pattern Recognition (CVPR)}, 2017.

\bibitem{Janner2017Self}
M.~Janner, J.~Wu, T.~D. Kulkarni, I.~Yildirim, and J.~Tenenbaum,
  ``Self-supervised intrinsic image decomposition,'' in \emph{Neural
  Information Processing Systems (NIPS)}, 2017.

\bibitem{Wu2017MarrNet}
J.~Wu, Y.~Wang, T.~Xue, X.~Sun, W.~T. Freeman, and J.~B. Tenenbaum, ``{MarrNet:
  3D Shape Reconstruction via 2.5D Sketches},'' in \emph{Neural Information
  Processing Systems (NIPS)}, 2017.

\bibitem{pix3d}
X.~Sun, J.~Wu, X.~Zhang, Z.~Zhang, C.~Zhang, T.~Xue, J.~B. Tenenbaum, and W.~T.
  Freeman, ``Pix3d: Dataset and methods for single-image 3d shape modeling,''
  in \emph{IEEE Conference on Computer Vision and Pattern Recognition (CVPR)},
  2018.

\bibitem{Allen1999Integration}
P.~K. Allen, A.~T. Miller, P.~Y. Oh, and B.~S. Leibowitz, ``Integration of
  vision, force and tactile sensing for grasping,'' \emph{Int. J. Intelligent
  Machines}, vol.~4, pp. 129--149, 1999.

\bibitem{Izatt2017Tracking}
G.~Izatt, G.~Mirano, E.~Adelson, and R.~Tedrake, ``Tracking objects with point
  clouds from vision and touch,'' in \emph{IEEE International Conference on
  Robotics and Automation (ICRA)}, 2017.

\bibitem{Bohg2010Strategies}
J.~Bohg, M.~Johnson-Roberson, M.~Bj{\"o}rkman, and D.~Kragic, ``Strategies for
  multi-modal scene exploration,'' in \emph{IEEE International Conference on
  Intelligent Robots and Systems (IROS)}, 2010.

\bibitem{Falco2017Cross}
P.~Falco, S.~Lu, A.~Cirillo, C.~Natale, S.~Pirozzi, and D.~Lee, ``Cross-modal
  visuo-tactile object recognition using robotic active exploration,'' in
  \emph{IEEE International Conference on Robotics and Automation (ICRA)}, 2017.

\bibitem{Luo2015Localizing}
S.~Luo, W.~Mou, K.~Althoefer, and H.~Liu, ``Localizing the object contact
  through matching tactile features with visual map,'' in \emph{IEEE
  International Conference on Robotics and Automation (ICRA)}, 2015.

\bibitem{Kroemer2011Learning}
O.~Kroemer, C.~H. Lampert, and J.~Peters, ``Learning dynamic tactile sensing
  with robust vision-based training,'' \emph{IEEE Trans. Robotics}, vol.~27,
  no.~3, pp. 545--557, 2011.

\bibitem{luo2018vitac}
S.~Luo, W.~Yuan, E.~Adelson, A.~G. Cohn, and R.~Fuentes, ``Vitac: Feature
  sharing between vision and tactile sensing for cloth texture recognition,''
  in \emph{IEEE International Conference on Robotics and Automation (ICRA)},
  2018.

\bibitem{Ottenhaus2016Local}
S.~Ottenhaus, M.~Miller, D.~Schiebener, N.~Vahrenkamp, and T.~Asfour, ``Local
  implicit surface estimation for haptic exploration,'' in \emph{IEEE-RAS
  International Conference on Humanoid Robots (Humanoids)}, 2016.

\bibitem{luo2016iterative}
S.~Luo, W.~Mou, K.~Althoefer, and H.~Liu, ``Iterative closest labeled point for
  tactile object shape recognition,'' in \emph{IEEE International Conference on
  Intelligent Robots and Systems (IROS)}, 2016.

\bibitem{Bierbaum2008Robust}
A.~Bierbaum, I.~Gubarev, and R.~Dillmann, ``Robust shape recovery for sparse
  contact location and normal data from haptic exploration,'' in \emph{IEEE
  International Conference on Intelligent Robots and Systems (IROS)}, 2008.

\bibitem{pezzementi2011object}
Z.~Pezzementi, C.~Reyda, and G.~D. Hager, ``Object mapping, recognition, and
  localization from tactile geometry,'' in \emph{IEEE International Conference
  on Robotics and Automation (ICRA)}, 2011.

\bibitem{Sommer2014Bimanual}
N.~Sommer, M.~Li, and A.~Billard, ``Bimanual compliant tactile exploration for
  grasping unknown objects,'' in \emph{IEEE International Conference on
  Robotics and Automation (ICRA)}, 2014.

\bibitem{Driess2017Active}
D.~Driess, P.~Englert, and M.~Toussaint, ``Active learning with query paths for
  tactile object shape exploration,'' in \emph{IEEE International Conference on
  Intelligent Robots and Systems (IROS)}, 2017.

\bibitem{Yi2016Active}
Z.~Yi, R.~Calandra, F.~Veiga, H.~van Hoof, T.~Hermans, Y.~Zhang, and J.~Peters,
  ``Active tactile object exploration with gaussian processes,'' in \emph{IEEE
  International Conference on Intelligent Robots and Systems (IROS)}, 2016.

\bibitem{Jamali2016Active}
N.~Jamali, C.~Ciliberto, L.~Rosasco, and L.~Natale, ``Active perception:
  Building objects' models using tactile exploration,'' in \emph{IEEE-RAS
  International Conference on Humanoid Robots (Humanoids)}, 2016.

\bibitem{martinez2013active}
U.~Martinez-Hernandez, G.~Metta, T.~J. Dodd, T.~J. Prescott, L.~Natale, and
  N.~F. Lepora, ``Active contour following to explore object shape with robot
  touch,'' in \emph{World Haptics Conference (WHC)}, 2013.

\bibitem{Luo2017Robotic}
S.~Luo, J.~Bimbo, R.~Dahiya, and H.~Liu, ``Robotic tactile perception of object
  properties: A review,'' \emph{Mechatronics}, vol.~48, pp. 54--67, 2017.

\bibitem{Varley2017Visual}
J.~Varley, D.~Watkins, and P.~Allen, ``Visual-tactile geometric reasoning,'' in
  \emph{RSS Workshop}, 2017.

\bibitem{Ilonen2014Three}
J.~Ilonen, J.~Bohg, and V.~Kyrki, ``Three-dimensional object reconstruction of
  symmetric objects by fusing visual and tactile sensing,'' \emph{Int. J.
  Robotics Res.}, vol.~33, no.~2, pp. 321--341, 2014.

\bibitem{Matsubara2017Active}
T.~Matsubara and K.~Shibata, ``Active tactile exploration with uncertainty and
  travel cost for fast shape estimation of unknown objects,'' \emph{Robotics
  Auton. Syst.}, vol.~91, pp. 314--326, 2017.

\bibitem{Yuan2017Connecting}
W.~Yuan, S.~Wang, S.~Dong, and E.~Adelson, ``Connecting look and feel:
  Associating the visual and tactile properties of physical materials,'' in
  \emph{IEEE Conference on Computer Vision and Pattern Recognition (CVPR)},
  2017.

\bibitem{Calandra2017Feeling}
R.~Calandra, A.~Owens, M.~Upadhyaya, W.~Yuan, J.~Lin, E.~H. Adelson, and
  S.~Levine, ``The feeling of success: Does touch sensing help predict grasp
  outcomes?'' in \emph{Conference on Robot Learning (CoRL)}, 2017.

\bibitem{He2015Deep}
K.~He, X.~Zhang, S.~Ren, and J.~Sun, ``Deep residual learning for image
  recognition,'' in \emph{IEEE Conference on Computer Vision and Pattern
  Recognition (CVPR)}, 2015.

\bibitem{dong2017improved}
S.~Dong, W.~Yuan, and E.~Adelson, ``Improved gelsight tactile sensor for
  measuring geometry and slip,'' in \emph{IEEE International Conference on
  Intelligent Robots and Systems (IROS)}, 2017.

\bibitem{Jakob2010Mitsuba}
W.~Jakob, ``Mitsuba renderer,'' 2010, http://www.mitsuba-renderer.org.

\bibitem{Xiao2010Sun}
J.~Xiao, J.~Hays, K.~A. Ehinger, A.~Oliva, and A.~Torralba, ``Sun database:
  Large-scale scene recognition from abbey to zoo,'' in \emph{IEEE Conference
  on Computer Vision and Pattern Recognition (CVPR)}, 2010.

\bibitem{Kingma2015Adam}
D.~P. Kingma and J.~Ba, ``Adam: A method for stochastic optimization,'' in
  \emph{International Conference on Learning Representations (ICLR)}, 2015.

\bibitem{jones2003non}
T.~R. Jones, F.~Durand, and M.~Desbrun, ``Non-iterative, feature-preserving
  mesh smoothing,'' in \emph{ACM Transactions on Graphics (TOG)}, vol.~22,
  no.~3.\hskip 1em plus 0.5em minus 0.4em\relax ACM, 2003, pp. 943--949.

\bibitem{Barrow1977Parametric}
H.~G. Barrow, J.~M. Tenenbaum, R.~C. Bolles, and H.~C. Wolf, ``Parametric
  correspondence and chamfer matching: Two new techniques for image matching,''
  in \emph{International Joint Conference on Artificial Intelligence (IJCAI)},
  1977.

\end{thebibliography}
